\documentclass{article}

\usepackage[final]{corl_2020} 

\newcommand{\etal}{\textit{et al.}}

\usepackage[dvipsnames]{xcolor}

\title{PLAS: Latent Action Space for Offline Reinforcement Learning}

\author{
  Wenxuan Zhou \qquad Sujay Bajracharya \qquad David Held\\
  Robotics Institute\\
  Carnegie Mellon University \\
  \texttt{\{wenxuanz, sbajrach, dheld\}@andrew.cmu.edu} \\
}

\usepackage[ruled,vlined]{algorithm2e}
\usepackage{enumitem}
\usepackage{wrapfig}
\usepackage{multirow}
\usepackage{graphicx}
\usepackage{amssymb}
\usepackage{bbm}
\usepackage{amsmath, bm}
\usepackage{float}

\usepackage[toc, page]{appendix}
\graphicspath{ {./figures/} }
\begin{document}
\maketitle
\begin{abstract}
    The goal of offline reinforcement learning is to learn a policy from a fixed dataset, without further interactions with the environment.  This setting will be an increasingly more important paradigm for real-world applications of reinforcement learning such as robotics, in which data collection is slow and potentially dangerous. Existing off-policy algorithms have limited performance on static datasets due to extrapolation errors from out-of-distribution actions. This leads to the challenge of constraining the policy to select actions within the support of the dataset during training. We propose to simply learn the \textbf{P}olicy in the \textbf{L}atent \textbf{A}ction \textbf{S}pace (PLAS) such that this requirement is naturally satisfied. We evaluate our method on continuous control benchmarks in simulation and a deformable object manipulation task with a physical robot. We demonstrate that our method provides competitive performance consistently across various continuous control tasks and different types of datasets, outperforming existing offline reinforcement learning methods with explicit constraints. Videos and code are available at \url{https://sites.google.com/view/latent-policy}.
\end{abstract}


\vspace{5mm}
\section{Introduction}

Reinforcement learning (RL) has achieved much success on many robotics tasks in simulation~\cite{human, deeploco}. 
However, it still has limited applications in the real world including real robots. One major challenge of applying RL in the real world is that it requires a large number of online interactions with the environment, usually more than millions of time steps. Offline Reinforcement Learning, or Batch Reinforcement Learning, aims to develop algorithms that can optimize the policy given a static dataset of transitions without any active data collection~\citep{offlineRL, batchRL}. This is especially important for robotics because algorithms that train from static datasets can provide additional flexibility in terms of data collection. We may take into account safety, use better exploration methods~\citep{curiosity}, and leverage demonstrations~\citep{Rajeswaran-RSS-18}. In addition, we can accumulate past experience during the development of the algorithm by re-using the replay buffer or evaluation trajectories from previous RL experiments. Furthermore, static datasets can be shared within the community, and thus, they are more likely to be scaled up in size. 

In contrast to offline RL, off-policy RL uses a replay buffer that stores transitions that are actively collected by the policy throughout a training procedure. 
Past work has shown that off-policy RL methods cannot be directly applied to static datasets due to the extrapolation error of the Q-function caused by out-of-distribution actions~\citep{BCQ}. To avoid extrapolation error, we need to constrain the policy to select actions \textit{within} the support of the dataset. On the other hand, the constraint cannot be ``overly restrictive"; in the extreme case, an overly constrained policy will degenerate to behavior cloning on the dataset. The design of such a constraint remains a challenging problem.

\begin{figure}
\centering
\includegraphics[width=\linewidth]{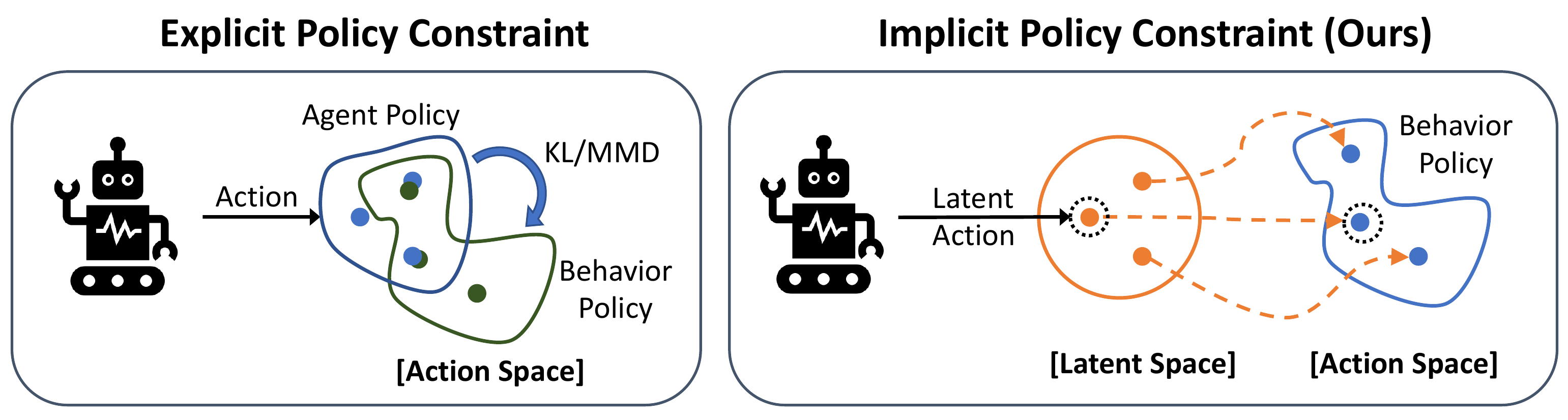}
\caption{Overview: Instead of explicitly matching the action distribution of the agent policy with the behavior policy using divergence metrics such as KL or MMD, we implicitly constrain the policy to output actions within the support of the behavior policy through the latent action space.}
\label{fig:overview}
\end{figure}

We propose a simple yet effective method that trains the Policy in the Latent Action Space (PLAS) to implicitly constrain the policy to output actions within the support of the dataset instead of using explicit constraints, as illustrated in Figure~\ref{fig:overview}. Following previous work, we model the ``behavior policy" of the dataset as a Conditional Variational Autoencoder (CVAE).  Our insight is that we can learn a policy in the latent action space of the CVAE and then use its decoder to output an action in the original action space of the environment. The latent action space \emph{implicitly} constrains the policy by construction. The benefit of such a constraint is that it can be naturally satisfied without affecting the optimization of the other components and without being restricted by the density of the behavior policy distribution. 

We demonstrate that PLAS allows generalization within the dataset and can provide consistently good performance for datasets with diverse actions. In cases where the Q-function generalizes well without significant extrapolation error, we augment our approach by allowing out-of-distribution actions in a controlled way to achieve better performance. This explicit separation of in-distribution generalization and out-of-distribution generalization allows the user fine-grained control over the generalization of the method.
We evaluate our method on the continuous control tasks from the d4rl benchmark datasets~\citep{d4rl} as well as real-robot experiments on deformable object manipulation and show superior performance to previous methods, despite the simplicity of our approach.

\section{Related Work}
\label{sec:related_work}
\textbf{Offline Reinforcement Learning: }
Offline reinforcement learning studies the problem of learning policies from static datasets without any active data collection~\citep{offlineRL,batchRL}. Recent work proposes different approaches in this direction~\citep{rem, awr, rwr, yu2020mopo, kidambi2020morel}. It has been empirically shown that the performance of off-policy algorithms drastically degrades when directly applied to static datasets due to out-of-distribution actions~\citep{BCQ}. Several papers propose to avoid out-of-distribution actions by enforcing constraints on the policy such as using a KL-divergence constraint or maximum mean discrepancy (MMD) constraint~\cite{Way,BEAR,BRAC}. In the most similar approach to our work, Fujimoto, \etal~\citep{BCQ} (BCQ) propose to learn a generative model for the behavior policy and perturb the randomly generated samples to find good perturbed actions that maximize the Q-function. Our experiments show that these approaches have worse performance than our method, likely due to the difficulty of satisfying the constraints or balancing in-distribution vs out-of-distribution generalization. 

\textbf{Imitation Learning:} The most naive way of using a static dataset is to perform behavior cloning. This approach is usually used when the dataset is generated by an expert policy. Behavior cloning only mimics the actions in the dataset and does not reason about which actions in the dataset are better than others. Imitation learning methods sometimes also assume access to an expert policy~\citep{dagger} and may allow interactive data in the environment~\citep{gail}, which is very different from offline RL.

\textbf{Generative Models for Actions:} Previous work has used a conditional variational autoencoder to model actions, although not in the offline RL setting. \citet{DBLP:journals/corr/MishraAM17} samples action sequences from the CVAE when they perform trajectory optimization with the learned latent dynamics model. ~\citet{Multi-agent} extended the previous method to multi-agent RL by learning a disentangled latent action representation. In contrast to these works, we focus on demonstrating the capability of using a CVAE over actions to deal with the out-of-distribution issue in off-policy RL in the offline setting.

\section{Background}
\subsection{Preliminaries}
\label{sec:preliminaries}

As is common in reinforcement learning, we define the environment as a Markov Decision Process (MDP) represented as the tuple $\mathcal{M}=(\mathcal{S},\mathcal{A},\mathcal{P},r,\gamma)$, where $\mathcal{S}$ is the state space, $\mathcal{A}$ is the action space, $\mathcal{P}:\mathcal{S} \times \mathcal{A} \times \mathcal{S} \rightarrow [0,1]$ is the transition probability function, $r: \mathcal{S} \times \mathcal{A} \times \mathcal{S} \rightarrow \mathbb{R}$ is the reward function, and $\gamma$ is the discount factor. The general objective of RL is to find a policy that maximizes the expectation of the return $\smash{G_t=\sum_{k=0}^{\infty} \gamma^{k}r(s_{t+k},a_{t+k},s_{t+k+1})}$.

Given a policy $\pi$, the action-value function, or Q-function, is defined as $Q^{\pi}(s,a)=\mathbb{E}_\pi[G_t|S_t=s, A_t=a]$. Our method builds on top of the commonly used off-policy actor-critic procedure with a deterministic policy~\citep{ddpg, td3}. The Q-function of the deterministic policy $\pi$ is estimated based on the Bellman Operator:
\vspace{-3mm}
\begin{equation}
\label{bellman}
\mathcal{T} \hat{Q}^{\pi}(s_t,a_t)=\mathbb{E}_{s_{t+1}}[r_t+\gamma \hat{Q}^{\pi}(s_{t+1}, \pi(s_{t+1}))]
\end{equation}
The policy $\pi_\theta$ is updated following the Deterministic Policy Gradient~\citep{DPG}:
\begin{equation}
\label{pg}
\nabla_{\theta}J(\theta)=\mathbb{E}_{s\sim \rho_{\pi}}[\nabla_{\theta}\pi_{\theta}(s) \nabla_a Q^{\pi_\theta}(s,a)|_{a=\pi_\theta (s)}] 
\end{equation}

\subsection{Offline RL: From Pessimistic MDP to Policy Constraints}
\label{sec:background}
In this section, we will discuss the objectives for offline reinforcement learning and the limitations of existing methods that build on top of off-policy RL.

In offline RL, we are given a fixed dataset $\mathcal{D}=\{(s_t,a_t,r_t,s_{t+1})_i\}$ with a finite number of transitions. The difficulty comes from the fact that the static dataset does not cover the entire state space and action space of the MDP. This is especially true when the state and action spaces are continuous. The objective of offline RL is typically to find the policy that maximizes the cumulative reward during its deployment in the environment. However, the performance of the policy will be limited by our knowledge over the MDP, which is inferred from a limited set of transitions. 

Reconsidering this problem, another reasonable objective for offline RL is to maximize the cumulative reward of the MDP under the transitions that have been visited in the dataset. Following~\citep{kidambi2020morel}, we may assume a pessimistic MDP such that $r(s,a)$ is significantly small for any unvisited $(s,a)$. Optimizing under such a pessimistic MDP is an intuitive surrogate objective. In addition, \citep{kidambi2020morel} proves that the performance of any policy for such a pessimistic MDP is a lower bound in the true MDP. 

An additional reason to optimize for this pessimistic MDP is the extrapolation error of approximated Q-functions~\citep{BCQ}. In off-policy algorithms, we bootstrap $Q(s_t,a_t)$ by using $Q(s_{t+1}, \pi(s_{t+1}))$ according to the Bellman operator as in Equation \eqref{bellman}. If $(s_{t+1}, \pi(s_{t+1}))$ is not in the dataset, $Q(s_{t+1}, \pi(s_{t+1}))$ can be arbitrarily wrong. This error caused by out-of-distribution \textit{actions} will be accumulated and exacerbated by the policy update. (Note that out-of-distribution \textit{states} do not occur during training.) Thus, optimizing the policy under the pessimistic MDP is equivalent to forcing the policy to select known actions that avoid any error accumulation. This is the motivation for constraining the policy to be within the support of the dataset. On the other hand, the constraint should not be overly restrictive and should not be affected by the distribution of the dataset as proposed in~\citep{BEAR}. The policy should have the full flexibility to choose actions within the support.

Existing offline RL methods enforce such a constraint in different ways. BCQ~\citep{BCQ} constrains the policy by sampling from the behavior policy. However, the policy is then restricted by the distribution of the behavior policy. BEAR~\citep{BEAR} and BRAC~\citep{BRAC} incorporate the constraint on the policy as a regularization term into the optimization process for the policy or the Q-function. This regularization term is calculated by a divergence metric, such as KL-divergence or MMD. There are two practical challenges to these approaches. First, this additional loss term creates a trade-off between optimizing the original objective and satisfying the constraint. Although BEAR uses a Lagrangian multiplier to solve  a constrained optimization problem, in practice, the constraint is almost never satisfied~\citep{BRAC}. Second, the choice of the divergence metric, such as KL and MMD, might be overly restrictive given datasets that have diverse actions. We provide further discussion on MMD constraint in Appendix \ref{ab:mmd}. These past approaches have shown that properly enforcing an explicit policy constraint is difficult; this observation motivates our approach. 

\subsection{Variational Auto-encoder}
Since our method uses a conditional variational autoencoder, we include a brief background of VAE in this section in its most general form based on~\cite{vae} and~\cite{vae_tutorial}. Given a dataset $X=\{x^{(i)}\}_{i=1}^N$, the goal of a VAE is to generate samples that are from the same distribution as the data points, in other words, to maximize $p(x)$ for all $x^{(i)}$. This is achieved by introducing a latent variable $z$ sampled from a prior distribution $p(z)$ and modeling a decoder $p_{\theta}(x|z)$ with parameter $\theta$. Directly maximizing the marginal likelihood $p_{\theta}(x)=\int p(z)p_{\theta}(x|z)dz$ is intractable. Instead,~\citet{vae} propose to approximate the true posterior $p_{\theta}(z|x)$ by training an encoder $q_{\phi}(z|x)$. In this way, they derive the following evidence lower bound (ELBO) on the log-likelihood of the data:

\vspace{-2mm}
\begin{equation}
\label{elbo-0}
    \max_{\theta, \phi}\log p(x) \ge \max_{\theta, \phi}\mathbb{E}_{q_\phi(z|x)}[\log p_\theta(x|z)]-\mathcal{D}_{KL}[q_\phi(z|x)||p_\theta(z)]
\end{equation}
\vspace{-3mm}

The term $\log p_\theta(x|z)$ (where $z$ is sampled from $q_\phi(z|x)$) represents the reconstruction loss. The second term is the KL-divergence between the encoder output and the prior of $z$, which is usually set to be $\mathcal{N}(0, 1)$. Thus, optimizing for this objective enables us to train a model that generates samples similar to the data distribution by sampling $z$ and then passing it into the decoder.
\section{Method}
\label{sec:method}

\begin{figure}
\centering
\vspace{-3mm}
\includegraphics[width=11cm]{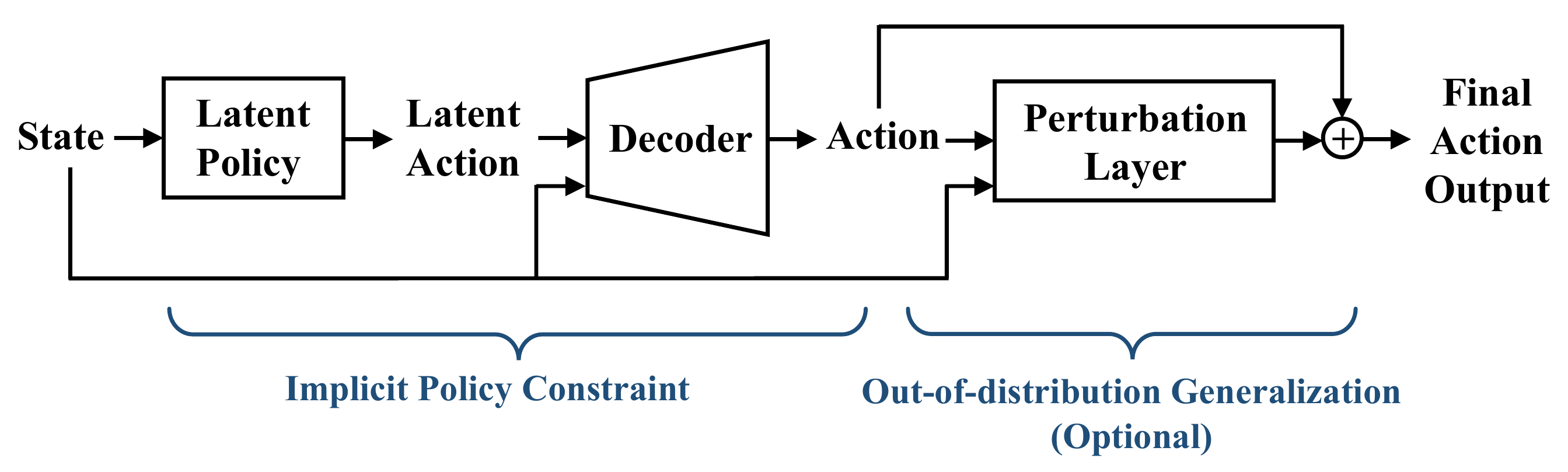}
\vspace{-3mm}
\caption{Network architecture for PLAS: Given a state, the latent policy outputs a latent action, which is then input into the decoder. The latent action space implicitly defines a constraint over the action output. An optional perturbation layer can be added on top of the output from the decoder to allow controlled generalization out of the training distribution. }
\vspace{-3mm}

\label{fig:method}
\end{figure}

In this section, we introduce our method PLAS (Policy with Latent Action Space) that implicitly constrains the policy to be within the support of the behavior policy. Our method  disentangles the in-distribution and out-of-distribution generalization of actions, enabling fine-grained control over the generalization of the method. The network architecture of our method is shown in Figure~\ref{fig:method}. 

\subsection{Policy in Latent Action Space (PLAS)}
\label{method:latent_policy}
Given a static dataset, we use a conditional variational autoencoder (CVAE) to model the behavior policy $p(a|s)$, as in other recent methods~\citep{BCQ, BEAR, BRAC}. The CVAE is trained to reconstruct actions conditioned on the states. Converting Equation \ref{elbo-0} into our problem formulation, the objective of the CVAE is to maximize $\log p(a|s)$ by maximizing its lower bound:
\begin{equation}
\label{elbo}
\max_{\alpha,\beta}\log p(a|s) \ge \max_{\alpha,\beta} \mathbb{E}_{z\sim q_\alpha}[\log p_\beta(a|s, z)]-\mathcal{D}_{KL}[q_\alpha(z|a,s)||P(z|s)]
\end{equation}
where $z$ is the latent variable, $\alpha$ and $\beta$ are the parameters of the encoder and the decoder, respectively. This is similar to Equation~\ref{elbo-0}, except that all terms are conditioned on the state $s$. A trained decoder $p_\beta(a|s, z)$ provides a mapping from the latent space to the action space, conditioned on the state. 

In order to constrain the policy to be within the support of the dataset, we propose to train a deterministic policy $z=\pi(s)$ to map from a state $s$ to a ``latent action" $z$; we then use the pretrained decoder $p_\beta(a|s, z)$ to project the latent action into the actual action space as shown in Figure~\ref{fig:method}. This is significantly different from BCQ which samples from a fixed range of the latent space. The latent policy has the flexibility of choosing the latent actions and thus avoids being affected by the density of the dataset distribution. The CVAE is trained to maximize $\log p(a|s)$, equivalent to maximizing the expected output of the decoder $\mathbb{E}_{p(z|s)}[p_\beta(a|s, z)]$ as shown:
\vspace{-4mm}

\begin{equation}
p(a|s)=\int_{z}p_\beta(a|s,z)p(z|s)dz=\mathbb{E}_{p(z|s)}[p_\beta(a|s, z)]
\label{eqn:decoder}
\end{equation}

Thus, for values of $z$ that have a high probability under the prior $p(z|s)$, the decoder $p_\beta(a|s, z)$ will output a high probability action under the behavior policy distribution $p(a|s)$ in expectation.

If we constrain the latent policy $z=\pi(s)$ to output a latent action $z$ that has a high probability under the prior $p(z|s)$, then the full policy, formed by $p_\beta(a|s, z=\pi(s))$, is likely to have a high probability under the behavior policy $p(a|s)$. Fortunately, this constraint is simple to enforce; since the prior $p(z|s)$ is set to a normal distribution $\mathcal{N}(0, 1)$, we simply define $z = \pi(s)$ such that $z_i \in [-\sigma, \sigma]$ for each latent dimension $i$ for some hyperparameter $\sigma$ (see Section~\ref{sec:Implementation Details} for details). 
Furthermore, for each state $s$, the latent policy has the flexibility to choose any latent action $z$ in this constrained latent action space, leading to an easier optimization compared to previous work with explicit constraints. 

\subsection{Generalization out of the dataset}
\label{sec:Generalization out of the dataset}
The latent policy provides a natural constraint to stay within the support of the dataset. However, in some cases when the Q-function can generalize well, we may relax the pessimistic objective and allow the policy to select out-of-distribution actions to improve its performance. The benefit of the out-of-distribution actions is more likely to happen when the environment (both transition probabilities and the reward function) is smooth and when the dataset has limited quality and diversity. To do so, we add a perturbation layer to the output of the decoder that outputs a residual over the action. The residual is limited to a specific range $[-\epsilon, \epsilon]$, where $\epsilon$ is a hyperparameter. Mathematically, this enforces the final output action to be close to the actions within the dataset in terms of the $L_{\infty}$ norm. This perturbation layer is inspired by BCQ. However, BCQ forms a policy by sampling from the generative model; the perturbation layer is used to prevent ``sampling from the generative model for a prohibitive number of times"~\citep{BCQ}. In our case, we don't perform any sampling since our policy is deterministic; the perturbation layer is instead specifically designed for out-of-distribution generalization. In the experiments, we will demonstrate that when the dataset has enough coverage in the state-action space, this additional layer is not necessary. 

\begin{algorithm}
\SetAlgoLined
 \textbf{Input}: Dataset $\mathcal{D}=\{(s_t,a_t,r_t,s_{t+1})_i\}$\\
 \textbf{// VAE Training}\\
 Initialize encoder $E_{\alpha}$ and decoder $D_{\beta}$ with parameters $\alpha$ and $\beta$.\\
 \For{$i \leftarrow \ 1$ \KwTo $M$}{
    Sample a minibatch of $k$ state-action pairs $(s_t,a_t)$ from $\mathcal{D}$\\
    Optimize $\alpha$ and $\beta$ using Equation \ref{elbo}\\
 }
 \textbf{// Policy Training}\\
 Initialize the latent policy network $\pi_{\theta}$, critic networks $Q_{\phi_1}$, $Q_{\phi_2}$ and their corresponding target networks $\pi_{\theta'}$, $Q_{\phi'_1}$ and $Q_{\phi'_2}$ with $\theta' \leftarrow \theta$, $\phi'_1 \leftarrow \phi_1$, $\phi_2' \leftarrow \phi_2$. \\
 \For{$i \leftarrow \ 1$ \KwTo $N$}{
    Sample a minibatch of $k$ transitions $\{ (s_t,a_t,r_t,s_{t+1})_{i=1,..,k} \}$ from $\mathcal{D}$\\
    For each transition, generate a latent action using the latent policy: $\psi_{t+1} = \pi_{\theta}(s_{t+1})$\\
    Decode latent actions using the decoder $a_{t+1}=D_{\beta}(\psi_{t+1})$\\
    Set $y= \lambda \min_{i=1, 2} Q'_{\phi_i}(s_{t+1}, a_{t+1}) + (1-\lambda) \max_{i=1, 2} Q'_{\phi_i}(s_{t+1}, a_{t+1})$ \\
    Update critic by minimizing: $L = (Q_{\phi_i}(s,a)-(r+\gamma y))^2$ for $i=1,2$\\
    Update actor according to Equation \ref{pg}\\
    Update target networks: $\theta' \leftarrow \tau \theta + (1-\tau) \theta'$, $\phi'_i \leftarrow \tau \phi_i + (1-\tau) \phi'_i$ for $i=1,2$. \\
 }
 \caption{Off-policy RL with PLAS}
 \label{algo}
\end{algorithm}
\vspace{-4mm}

\subsection{Implementation Details}
\label{sec:Implementation Details}
The full algorithm is summarized in Algorithm \ref{algo}. Our algorithm can be built on top of off-policy algorithm such as DDPG~\cite{ddpg} or TD3~\cite{td3}. We use a deterministic policy $z = \pi(s)$ to output a latent action. The policy uses a tanh activation at the output layer to limit the max latent action. This limit is set to 2 by default, which corresponds to $2\sigma$ for the latent variable $p(z)=\mathcal{N}(0, 1)$. We use a soft Clipped Double Q-learning with parameter $\lambda$ to weight the two Q-functions. Following common practice, we use target networks to stabilize training with hyperparameter $\tau$. The code for our algorithm is based on the BCQ repository, and we mostly follow the hyperparameters of BCQ. Further implementation details and hyperparameters can be found in Appendix~\ref{ab:implementation}.
\section{Experiments}
\label{sec:result}

We evaluate our algorithm on a wide range of continuous control tasks, including a physical robot experiment on deformable object manipulation and the d4rl benchmarks~\citep{d4rl} including OpenAI Gym locomotion tasks, Adroit, Franka Kitchen, etc. For the d4rl benchmarks, our analysis in the main text is focused on the locomotion datasets; the full results including the other environments can be found in Appendix~\ref{ab:d4rl}. We compare our method with the following baselines: BCQ~\citep{BCQ}, BEAR~\citep{BEAR}, and BRAC~\citep{BRAC}. We use the author's implementation of these algorithms with recommended hyperparameters reported in these papers. Note that we use the latent policy without the perturbation layer by default because our primary focus is on the policy constraint and in-distribution generalization. The experiments that use the perturbation layer are explicitly mentioned.

\subsection{Experiment Descriptions}
\textbf{Real-Robot Experiment}: The task for the real-robot experiment is to slide along the edge of the cloth as far as possible with a tactile sensor. The experiment setup is shown in Figure~\ref{fig:robot}(a) and an example of the tactile sensor reading is shown in Figure~\ref{fig:robot}(b).
More details of this experiment can be found in Appendix~\ref{ab:robot}. The dataset consists of the replay buffer from a previous online RL experiment with around 7000 timesteps of transitions and 5 episodes (around 300 timesteps) of expert demonstrations from a trained policy.  We train the policy for 2400 steps in total for each experiment with evaluations every 150 steps over 5 episodes. 

\textbf{Locomotion Datasets}: We mainly focus on the locomotion environments from the d4rl datasets in this section including Walker2d-v2, Hopper-v2, and Halfcheetah-v2. For each environment, there are four types of datasets: random, medium, medium expert, and medium replay datasets. Random datasets are generated by randomly initialized policies. Medium datasets are generated from rollouts of a ``medium" performance policy trained with Soft Actor-Critic up to a certain performance. Medium-expert datasets are generated by combining the medium datasets and expert datasets. Medium-replay datasets are the replay buffers created during the training of the medium policies. Note that medium-replay datasets are much smaller than the other types of datasets, making it more challenging to obtain stable training performance. We train the policy for 500 epochs and each epoch has 1000 training steps. The policy is evaluated every 1 epoch over 10 episodes.

\subsection{Performance on Real-Robot Experiment}

\begin{figure}
\centering
\includegraphics[width=\linewidth]{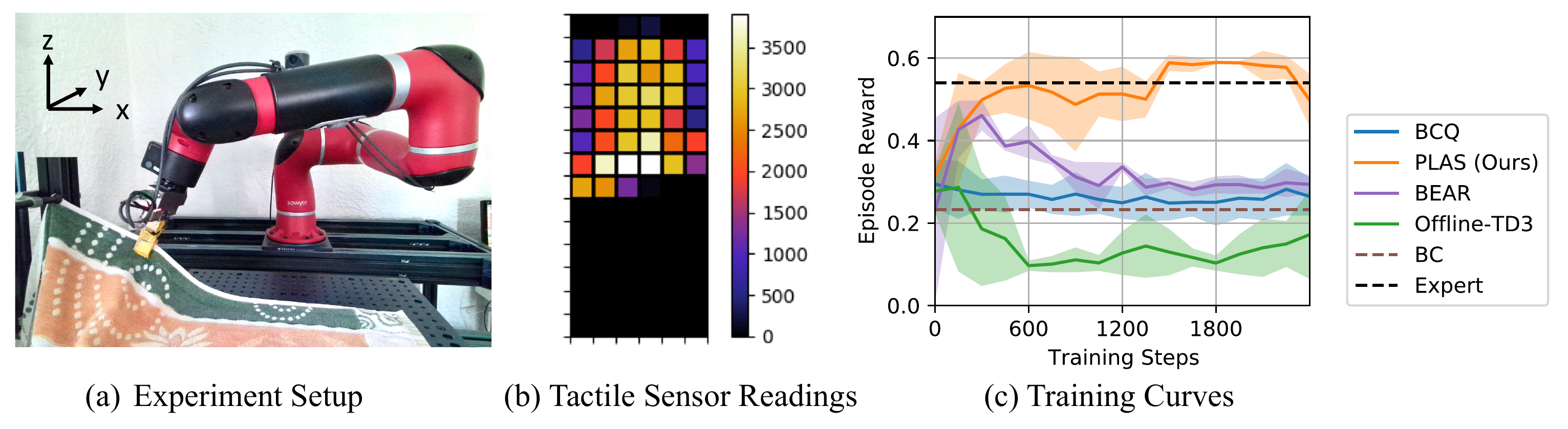}
\vspace{-5mm}
\caption{Real-robot experiment: (a) Experiment setup for the cloth sliding task. The cloth is fixed at the top left corner. (b) An example of the tactile sensor readings when the robot grasps at the edge of the cloth. (c) Training curves for the cloth sliding task on our method and the baselines. It shows the episode reward over five evaluation episodes every 150 training steps.}
\label{fig:robot}
\vspace{-2mm}
\end{figure}

Figure~\ref{fig:robot}(c) shows the evaluation performance of different methods across the training process for the cloth sliding task. The brown dashed line indicates the behavior cloning (BC) policy. It fails as expected because the average quality of the dataset is poor. The ``Offline-TD3" baseline is to directly run TD3 over the offline dataset without any extra online data collection. The performance is even worse than BC, which shows the necessity of designing offline RL algorithms that can utilize fixed datasets for real-world applications. BCQ also doesn't perform well, possibly because it is overly constrained by the dataset distribution, resulting in similar performance as behavior cloning. BEAR achieves reasonable performance at the beginning of training but it drops soon after. Our method outperforms all the baselines, and the final performance is similar to the expert policy. 

\subsection{Performance on D4RL datasets}

\begin{figure}
\centering
\includegraphics[width=\linewidth]{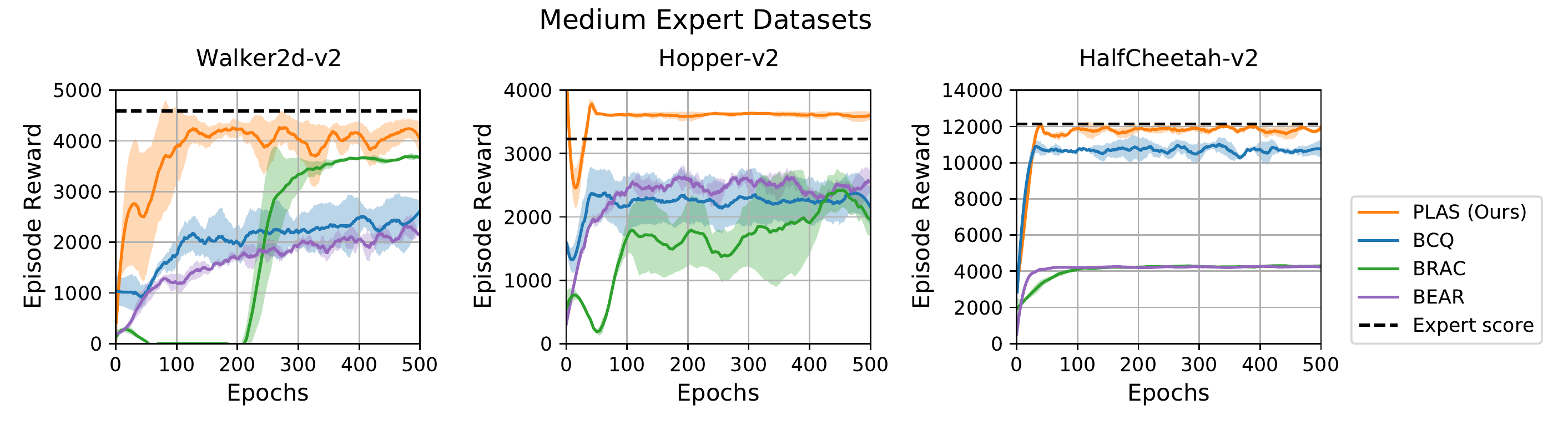}
\includegraphics[width=\linewidth]{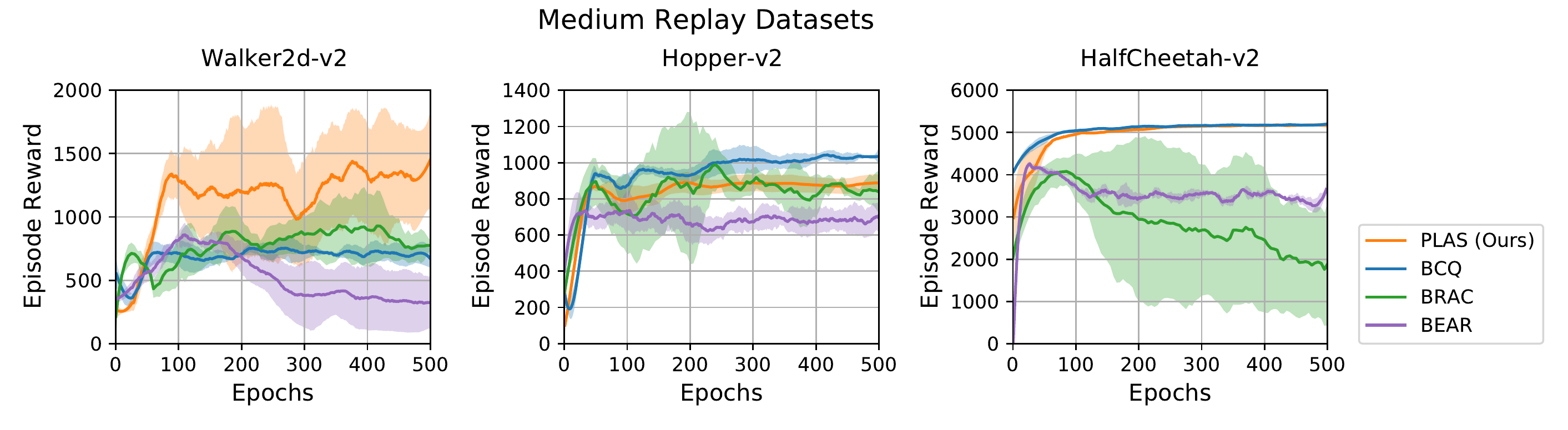}
\vspace{-5mm}
\caption{Training performance for medium-expert and medium-replay datasets on locomotion tasks. Each curve is averaged over 3 seeds. Shaded area shows one standard deviation across seeds.}
\label{fig:result}
\end{figure}

To more systematically benchmark the performance of our method with the other offline RL algorithms, we ran experiments on the d4rl benchmarks. We focus the discussions and analysis on the locomotion environments here; full results on the d4rl datasets can be found in Appendix~\ref{ab:d4rl} including Locomotion, Maze2d, AntMaze, Adroit Hand, Franka Kitchen environments. To highlight the performance of our method over the medium-expert datasets and the medium-replay datasets, we include the training curves in Figure~\ref{fig:result}. These two types of datasets are especially important because they have diverse coverage over states and actions generated by a mixture of policies. These kinds of diverse datasets are also more likely to appear in real-world applications with data collected from different sources. The diversity allows the potential to learn a good policy from the data; on the other hand, diversity also introduces difficulties for the policy constraints. Figure~\ref{fig:result} shows that we consistently achieve performance that is similar to or better than the best baselines on these datasets, demonstrating the effectiveness of our method in fully utilizing the datasets.

\subsection{Overestimation of Learned Q-functions}

\begin{figure}
\centering
\includegraphics[width=\linewidth]{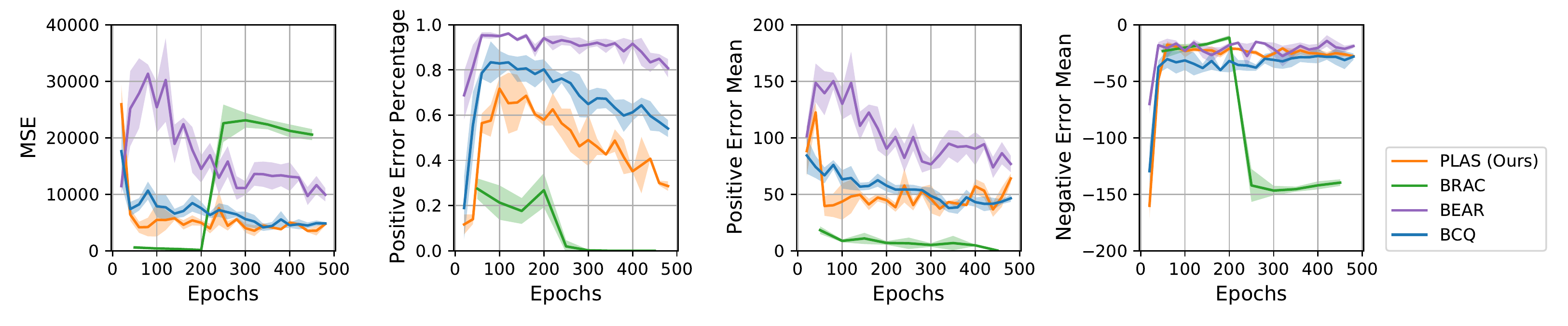}
\vspace{-5mm}
\caption{We perform an analysis of Q-function errors of different methods, using the following metrics: (a) Mean-squared error of the Q-values (b) The percentage of overestimated Q-values (c) Mean of the positive errors (magnitude of overestimation) (d) Mean of the negative errors (magnitude of underestimation)}
\label{fig:overestimation}
\end{figure}

We analyze the quality of the learned Q-function in detail with the Walker2d medium-expert dataset for different algorithms. By definition, the Q-value $Q_\pi(s_t, a_t)$ is equal to the expected return starting from state $s_t$ following action $a_t$; our learned Q-function attempts to estimate this value. Thus, we evaluate the Q-values by comparing them to the true returns for the transitions during rollouts. The true return is calculated by the cumulative discounted reward until termination or up to $r_{t+1000}$, since the reward after 1000 steps is negligible due to the discount factor. We define the estimation error to be $Q(s_t, a_t)-G(s_t, a_t)$, where $G(s_t, a_t)$ is the empirical return. Positive error corresponds to overestimation bias and negative error corresponds to underestimation bias. In Figure~\ref{fig:overestimation}, we show four metrics on the quality of the Q-function over $N$ transitions from 50 episodes: 
\begin{itemize}[leftmargin=*]
    \item \textbf{Mean Squared Error (MSE)} - measures the overall quality: $\sum_i(Q(s_i, a_i)-G(s_i, a_i))^2$) 
    \item \textbf{Positive Error Percentage} - percentage of overestimation: $\frac{1}{N}\sum_i(\mathbbm{1}(Q(s_i, a_i)-G(s_i, a_i))>0))$
    \item \textbf{Positive Error Mean} - the mean value of the positive errors, which indicates the average magnitude of over-estimation: Average of $Q(s_i, a_i)-G(s_i, a_i)$ for $Q(s_i, a_i)-G(s_i, a_i)>0$
    \item \textbf{Negative Error Mean} - the mean value of the negative errors, which indicates the average magnitude of under-estimation: Average of $Q(s_i, a_i)-G(s_i, a_i)$ for $Q(s_i, a_i)-G(s_i, a_i)<0$
\end{itemize}

MSE measures the overall estimation bias, and the other three metrics capture the direction of the bias. Our method achieves consistently low MSE during training compared with the baselines. Note that BRAC has low MSE during the beginning of training because the return is close to 0, as shown in the training curves in Figure~\ref{fig:result}. Although both our method and BRAC achieve similar performance at the end of training on the Walker2d medium-expert dataset (though our method converges faster), MSE indicates that our method results in a better Q-function. In terms of the direction of the bias, BEAR has a large overestimation bias and BRAC has a large underestimation bias. Overestimation bias is usually considered more harmful than underestimation for Q-learning based algorithms~\cite{td3}. As a result, although BRAC has a higher MSE than BEAR, the evaluation performance is still better. Our method does not have significant underestimation or overestimation in this case.

\subsection{Effect of the Optional Perturbation Layer}
\begin{table}
  \caption{Comparison of different perturbation values on random and medium datasets. Scores are normalized. $\epsilon=0$ is the performance of the latent policy without the additional perturbation layer.}
  \label{tab:perturbation}
  \centering
  \begin{tabular}{lcccccc}
    \hline\multicolumn{1}{l}{\bf Dataset}
    &\multicolumn{1}{c}{\bf $\epsilon=0$}
    &\multicolumn{1}{c}{\bf $\epsilon=0.05$}
    &\multicolumn{1}{c}{\bf $\epsilon=0.1$}
    &\multicolumn{1}{c}{\bf $\epsilon=0.2$}
    &\multicolumn{1}{c}{\bf $\epsilon=0.5$}\\\hline
    walker2d-random	        & $3.1$	        & \bm{$6.8$}	        & $2.4$     	& $1.3$	        & $-0.3$     \\
    hopper-random	        & $10.5$	    & $11.1$	    & $11.6$	    & $12.2$    	& \bm{$13.3$}  \\
    halfcheetah-random      & $25.8$	    & $25.7$	    & $27.4$ 	    & $27.6$	    & \bm{$28.3$}  \\
    walker2d-medium	        & $44.6$	    & $64.8$	    & \bm{$66.9$}	    & $62.1$	    & $39.2$   \\
    hopper-medium	        & $32.9$	    & \bm{$35.5$}	    & $17.5$	    & $2.5$	        & $2.1$   \\
    halfcheetah-medium	    & $39.3$	    & $41.3$    	& \bm{$42.2$}	    & \bm{$42.2$}	    & $40.4$  \\\hline

  \end{tabular}
\end{table}

In Section~\ref{sec:background}, we mentioned that when the Q-function generalizes well, allowing the policy to select some out-of-distribution actions might be helpful. This motivates us to introduce an additional perturbation layer as mentioned in Section~\ref{sec:Generalization out of the dataset}. We evaluate the benefit of this optional perturbation layer with different max perturbation limits, with $\epsilon=0$ being the latent policy alone. The results for a selective set of environments are summarized in Table~\ref{tab:perturbation}. Note that the action space in these tasks is defined to be $(-1,1)$; thus $\epsilon=0.5$ allows a very high range of perturbation. We found that the importance of the perturbation layer depends on both the dataset and the environment. Allowing out-of-distribution actions often leads to improved performance for random datasets. The ``medium" datasets tend to have peak performance with smaller values of $\epsilon$; a larger value likely leads to errors in the Q-function evaluation due to out-of-distribution state-action pairs. The full results including medium-expert and medium-replay datasets are in Appendix~\ref{ab:perturbation}. We found that medium-expert and medium-replay datasets usually do not benefit from the perturbation layer.

\section{Conclusion}
\label{sec:conclusion}

We propose a straightforward approach to offline RL that implicitly constrains the policy to be within the support of the dataset without being restricted by the density of the dataset distribution. Furthermore, we study the effect of an additional perturbation layer that allows out-of-distribution generalization of Q-functions. We demonstrate that our approach can effectively learn a policy with real-world data in the cloth sliding experiment and achieves competitive performance over offline RL benchmarks. By enabling a more efficient use of data from various sources, PLAS paves the way for future possibilities of using RL on real robots.


\clearpage
\acknowledgments{We thank Siddharth Ancha for insightful discussions. This material is based upon work supported by the United States Air Force and DARPA under Contract No. FA8750-18-C-0092, LG Electronics and the National Science Foundation under Grant No. IIS-1849154. }

\bibliography{example}  

\newpage
\begin{appendices}

\section{Implementation Details}
\label{ab:implementation}
The implementation of our algorithm is based on the original implementation of BCQ: \url{https://github.com/sfujim/BCQ}. We train the CVAE first and then train the policy using the fixed decoder. The latent policy is a deterministic policy with tanh activation at the output. The output is then scaled by a hyperparameter max latent action. More discussions on the max latent action is in Appendix \ref{ab:ma}. The perturbation layer is not used by default. We include the discussion on the effect of the perturbation layer in Appendix \ref{ab:perturbation}.

\textbf{Hyper-parameters for MuJoCo datasets}: The actor, the critic, and the CVAE are optimized using Adam. The actor learning rate is 1e-4 and the critic learning rate is 1e-3. The CVAE learning rate is 1e-4. Both the encoder and the decoder have two hidden layers (750, 750) by default. For datasets smaller than 1e6 transitions such as the medium-replay datasets, we use (128, 128) to prevent overfitting. We train the CVAE for 5e5 timesteps with batch size 100. The latent policy, the critic, and the perturbation layer have two hidden layers (400, 300). We use $\tau=0.005$ for the soft target update. $\lambda=1$ is used to calculate the Q-value target. The policy is trained for 5e5 timesteps with batch size 100.

\textbf{Hyper-parameters for the robot experiment}: 
The actor and critic learning rates are set to 3e-4 and the CVAE learning rate is 1e-4, with Adam as the optimizer. All of the networks have two hidden layers of size 64, including the actor, the critic, the encoder, and the decoder. The smaller network sizes are to prevent overfitting. The CVAE is trained for 15000 iterations. For soft target update we use $\tau=0.005$. We use $\lambda=0.75$ for clipped double Q learning and use batch size of 256. The max latent action is set to 2.0. 

\newpage
\section{D4RL Results}
\label{ab:d4rl}

To benchmark the performance of our algorithm, we include the full results for the d4rl MuJoCo datasets here as a reference. The numbers for the baselines are from the d4rl paper~\citep{d4rl}. The results are averaged over 3 seeds. ``PLAS" refers to the latent policy without the perturbation layer. ``PLAS+P" refers to the latent policy + perturbation layer. For max latent action, hopper-medium-replay and halfcheetah-medium-expert use 0.5 and all the other locomotion datasets use 2 for both ``PLAS" and ``PLAS+P". For the max perturbation, we report the best result from Appendix~\ref{ab:perturbation}. Our method consistently achieves good performance especially on medium-expert and medium-replay datasets. The other baselines work well on a part of the datasets and fail on the others.

In the current version of the d4rl dataset, hopper-medium-expert is actually a combination of the medium-replay and the expert datasets instead of the medium and the expert datasets. We have verified that the results given in their paper also correspond to the medium-replay + expert dataset. In Table \ref{tab:d4rl1} and Table \ref{tab:d4rl2} below, we use hopper-medium-expert(a) to refer to the results on this dataset. In addition, we generate the actual hopper-medium-expert by concatenating the medium and the expert datasets, referred to as hopper-medium-expert(b) in the table. The results from Figure \ref{fig:result} and the other experiments in the appendix are all based on hopper-medium-expert(b).

\begin{table}[H]
  \caption{D4rl Benchmark Results: Average Reward}
  \label{tab:d4rl1}
  \centering
  \begin{tabular}{lrrrrrr}
    \hline
    \multicolumn{1}{l}{\bf Dataset}
    &\multicolumn{1}{c}{\bf BEAR}
    &\multicolumn{1}{c}{\bf BRAC-v}
    &\multicolumn{1}{c}{\bf BCQ}
    &\multicolumn{1}{c}{\bf \begin{tabular}{@{}c@{}}PLAS \\ (Ours)\end{tabular}}
    &\multicolumn{1}{c}{\bf \begin{tabular}{@{}c@{}}PLAS+P \\ (Ours)\end{tabular}}\\\hline
    walker2d-medium-expert	    & $1842.7$	&$4926.6$	&$2640.3$   &$4113.2$	&$4465.0$\\
    hopper-medium-expert-(a)	& $3113.5$	&$5.1$	    &$3588.5$   &$3593.7$	&$3062.5$\\
    hopper-medium-expert-(b)	& $2648.4$	&$2245.7$	&$2021.7$   &$3592.4$	&$3518.5$\\
    halfcheetah-medium-expert   & $6349.6$	&$4926.6$	&$7750.8$   &$11716.9$   &$12051.4$\\\hline
    walker2d-medium-replay	    & $883.8$	&$44.5$	    &$688.7$    &$1387.9$	&$658.4$\\
    hopper-medium-replay	    & $1076.8$	&$-0.8$	    &$1057.8$   &$888.4$	&$1669.6$\\
    halfcheetah-medium-replay	& $4517.9$	&$5640.6$	&$4463.9$   &$5172.6$   &$5397.4$\\\hline
    walker2d-medium	            & $2717.0$	&$3725.8$	&$2441.0$   &$2047.0$	&$3072.4$\\
    hopper-medium	            & $1674.5$	&$990.4$	&$1752.4$   &$1050.4$	&$1182.1$\\
    halfcheetah-medium	        & $4897.0$	&$5473.8$	&$4767.9$   &$4602.6$   &$4964.6$\\\hline
    walker2d-random	            & $336.3$	&$87.4$	    &$228.0$    &$104.0$	&$311.6$\\
    hopper-random	            & $349.9$	&$376.3$	&$323.9$    &$320.5$	&$412.2$\\
    halfcheetah-random	        & $2831.4$	&$3590.1$	&$-1.3$     &$2922$     &$3235.8$\\\hline

  \end{tabular}
\end{table}

\begin{table}[H]
  \caption{D4rl Benchmark Results: Normalized Score}
  \label{tab:d4rl2}
  \centering
  \begin{tabular}{lrrrrrr}
    \hline\multicolumn{1}{l}{\bf Dataset}
    &\multicolumn{1}{c}{\bf BEAR}
    &\multicolumn{1}{c}{\bf BRAC-v}
    &\multicolumn{1}{c}{\bf BCQ}
    &\multicolumn{1}{c}{\bf \begin{tabular}{@{}c@{}}PLAS \\ (Ours)\end{tabular}}
    &\multicolumn{1}{c}{\bf \begin{tabular}{@{}c@{}}PLAS+P \\ (Ours)\end{tabular}}\\\hline
    walker2d-medium-expert	& $40.1$	&$81.6$	&$57.5$	&$89.6$	&$97.2$\\
    hopper-medium-expert-(a)	& $96.3$	&$0.8$	&$110.9$	&$111.0$	&$94.7$\\
    hopper-medium-expert-(b)	& $82.0$	&$69.6$	&$62.7$	&$111.0$	&$108.7$\\
    halfcheetah-medium-expert   & $53.4$	&$41.9$	&$64.7$	&$96.6$	&$99.3$\\\hline
    walker2d-medium-replay	& $19.2$	&$0.9$	&$15$	&$30.2$	&$14.3$\\
    hopper-medium-replay	& $33.7$	&$0.6$	&$33.1$	&$27.9$	&$51.9$\\
    halfcheetah-medium-replay	& $38.6$	&$47.7$	&$38.2$	&$43.9$	&$45.7$\\\hline
    walker2d-medium	& $59.1$	&$81.1$	&$53.1$	&$44.6$	&$66.9$\\
    hopper-medium	& $52.1$	&$31.1$	&$54.5$	&$32.9$	&$36.9$\\
    halfcheetah-medium	& $41.7$	&$46.3$	&$40.7$	&$39.3$	&$42.2$\\\hline
    walker2d-random	& $7.3$	&$1.9$	&$4.9$	&$3.1$	&$6.8$\\
    hopper-random	& $11.4$	&$12.2$	&$10.6$	&$10.5$	&$13.3$\\
    halfcheetah-random	& $25.1$	&$31.2$	&$2.2$	&$25.8$	&$28.3$\\\hline

  \end{tabular}
\end{table}

\begin{table}[H]
  \caption{D4rl Results on More Datasets: Average Reward. For these datasets, we searched over $0.5, 1, 2$ for max latent action and report the best results.}
  \label{tab:d4rl3}
  \centering
  \begin{tabular}{lrrrrrr}
    \hline\multicolumn{1}{l}{\bf Dataset}
    &\multicolumn{1}{c}{\bf BC}
    &\multicolumn{1}{c}{\bf SAC-off}
    &\multicolumn{1}{c}{\bf BEAR}
    &\multicolumn{1}{c}{\bf BRAC-v}
    &\multicolumn{1}{c}{\bf BCQ}
    &\multicolumn{1}{c}{\bf \begin{tabular}{@{}c@{}}PLAS \\ (Ours)\end{tabular}}\\\hline
    
    maze2d-umaze	& $	29.0	$ & $	145.6	$ & $	28.6	$ & $	1.7	$ & $	41.5	$ & $	102.6	$\\
    maze2d-medium	& $	93.2	$ & $	82.0	$ & $	89.8	$ & $	102.4	$ & $	35.0	$ & $	109.6	$\\
    maze2d-large	& $	20.1	$ & $	1.5	$ & $	19.0	$ & $	115.2	$ & $	23.2	$ & $	334.6	$\\ \hline
    													
    antmaze-umaze	& $	0.7	$ & $	0.0	$ & $	0.7	$ & $	0.7	$ & $	0.6	$ & $	0.7	$\\
    antmaze-umaze-diverse	& $	0.6	$ & $	0.0	$ & $	0.6	$ & $	0.7	$ & $	0.7	$ & $	0.5	$\\
    antmaze-medium-play	& $	0.0	$ & $	0.0	$ & $	0.0	$ & $	0.0	$ & $	0.0	$ & $	0.2	$\\
    antmaze-medium-diverse	& $	0.0	$ & $	0.0	$ & $	0.1	$ & $	0.0	$ & $	0.0	$ & $	0.0	$\\
    antmaze-large-play	& $	0.0	$ & $	0.0	$ & $	0.0	$ & $	0.0	$ & $	0.0	$ & $	0.0	$\\
    antmaze-large-diverse	& $	0.0	$ & $	0.0	$ & $	0.0	$ & $	0.0	$ & $	0.0	$ & $	0.0	$\\ \hline
    													
    pen-human	& $	1121.9	$ & $	284.8	$ & $	66.3	$ & $	114.7	$ & $	2149.0	$ & $	2101.0	$\\
    hammer-human	& $	-82.4	$ & $	-214.2	$ & $	-242.0	$ & $	-243.8	$ & $	-210.5	$ & $	324.7	$\\
    door-human	& $	-41.7	$ & $	57.2	$ & $	-66.4	$ & $	-66.4	$ & $	-56.6	$ & $	73.3	$\\
    relocate-human	& $	-5.6	$ & $	-4.5	$ & $	-18.9	$ & $	-19.7	$ & $	-8.6	$ & $	7.1	$\\
    pen-cloned	& $	1791.8	$ & $	797.6	$ & $	885.4	$ & $	22.2	$ & $	1407.8	$ & $	1558.0	$\\
    hammer-cloned	& $	-175.1	$ & $	-244.1	$ & $	-241.1	$ & $	-236.9	$ & $	-224.4	$ & $	-142.9	$\\
    door-cloned	& $	-60.7	$ & $	-56.3	$ & $	-60.9	$ & $	-59.0	$ & $	-56.3	$ & $	41.2	$\\
    relocate-cloned	& $	-10.1	$ & $	-16.1	$ & $	-17.6	$ & $	-19.4	$ & $	-17.5	$ & $	-16.7	$\\
    pen-expert	& $	2633.7	$ & $	277.4	$ & $	3253.1	$ & $	6.4	$ & $	3521.3	$ & $	3693.3	$\\
    hammer-expert	& $	16140.8	$ & $	3019.5	$ & $	16359.7	$ & $	-241.1	$ & $	13731.5	$ & $	16333.5	$\\
    door-expert	& $	969.4	$ & $	163.8	$ & $	2980.1	$ & $	-66.6	$ & $	2850.7	$ & $	3004.0	$\\
    relocate-expert	& $	4289.3	$ & $	-18.2	$ & $	4173.8	$ & $	-21.4	$ & $	1759.6	$ & $	4528.5	$\\ \hline
    													
    kitchen-complete	& $	1.4	$ & $	0.6	$ & $	0.0	$ & $	0.0	$ & $	0.3	$ & $	1.4	$\\
    kitchen-partial	& $	1.4	$ & $	0.0	$ & $	0.5	$ & $	0.0	$ & $	0.8	$ & $	1.8	$\\
    kitchen-mixed	& $	1.9	$ & $	0.1	$ & $	1.9	$ & $	0.0	$ & $	0.3	$ & $	1.6	$\\ \hline

  \end{tabular}
\end{table}

\begin{table}[H]
  \caption{D4rl Results on More Datasets: Normalized Score}
  \label{tab:d4rl4}
  \centering
  \begin{tabular}{lrrrrrr}
    \hline\multicolumn{1}{l}{\bf Dataset}
    &\multicolumn{1}{c}{\bf BC}
    &\multicolumn{1}{c}{\bf SAC-off}
    &\multicolumn{1}{c}{\bf BEAR}
    &\multicolumn{1}{c}{\bf BRAC-v}
    &\multicolumn{1}{c}{\bf BCQ}
    &\multicolumn{1}{c}{\bf \begin{tabular}{@{}c@{}}PLAS \\ (Ours)\end{tabular}}\\\hline
    
    maze2d-umaze	& $	3.8	$ & $	88.2	$ & $	3.4	$ & $	-16.0	$ & $	12.8	$ & $	57.0	$\\
    maze2d-medium	& $	30.3	$ & $	26.1	$ & $	29.0	$ & $	33.8	$ & $	8.3	$ & $	36.5	$\\
    maze2d-large	& $	5.0	$ & $	-1.9	$ & $	4.6	$ & $	40.6	$ & $	6.2	$ & $	122.7	$\\ \hline
    													
    antmaze-umaze	& $	65.0	$ & $	0.0	$ & $	73.0	$ & $	70.0	$ & $	78.9	$ & $	70.7	$\\
    antmaze-umaze-diverse	& $	55.0	$ & $	0.0	$ & $	61.0	$ & $	70.0	$ & $	55.0	$ & $	45.3	$\\
    antmaze-medium-play	& $	0.0	$ & $	0.0	$ & $	0.0	$ & $	0.0	$ & $	0.0	$ & $	16.0	$\\
    antmaze-medium-diverse	& $	0.0	$ & $	0.0	$ & $	8.0	$ & $	0.0	$ & $	0.0	$ & $	0.7	$\\
    antmaze-large-play	& $	0.0	$ & $	0.0	$ & $	0.0	$ & $	0.0	$ & $	6.7	$ & $	0.7	$\\
    antmaze-large-diverse	& $	0.0	$ & $	0.0	$ & $	0.0	$ & $	0.0	$ & $	2.2	$ & $	0.3	$\\ \hline
    													
    pen-human	& $	34.4	$ & $	6.3	$ & $	-1.0	$ & $	0.6	$ & $	68.9	$ & $	67.3	$\\
    hammer-human	& $	1.5	$ & $	0.5	$ & $	0.3	$ & $	0.2	$ & $	0.5	$ & $	4.6	$\\
    door-human	& $	0.5	$ & $	3.9	$ & $	-0.3	$ & $	-0.3	$ & $	0.0	$ & $	4.4	$\\
    relocate-human	& $	0.0	$ & $	0.0	$ & $	-0.3	$ & $	-0.3	$ & $	-0.1	$ & $	0.3	$\\
    pen-cloned	& $	56.9	$ & $	23.5	$ & $	26.5	$ & $	-2.5	$ & $	44.0	$ & $	49.0	$\\
    hammer-cloned	& $	0.8	$ & $	0.2	$ & $	0.3	$ & $	0.3	$ & $	0.4	$ & $	1.0	$\\
    door-cloned	& $	-0.1	$ & $	0.0	$ & $	-0.1	$ & $	-0.1	$ & $	0.0	$ & $	3.3	$\\
    relocate-cloned	& $	-0.1	$ & $	-0.2	$ & $	-0.3	$ & $	-0.3	$ & $	-0.3	$ & $	-0.2	$\\
    pen-expert	& $	85.1	$ & $	6.1	$ & $	105.9	$ & $	-3.0	$ & $	114.9	$ & $	120.7	$\\
    hammer-expert	& $	125.6	$ & $	25.2	$ & $	127.3	$ & $	0.3	$ & $	107.2	$ & $	127.1	$\\
    door-expert	& $	34.9	$ & $	7.5	$ & $	103.4	$ & $	-0.3	$ & $	99.0	$ & $	104.2	$\\
    relocate-expert	& $	101.3	$ & $	-0.3	$ & $	98.6	$ & $	-0.4	$ & $	41.6	$ & $	106.9	$\\ \hline
    													
    kitchen-complete	& $	33.8	$ & $	15.0	$ & $	0.0	$ & $	0.0	$ & $	8.1	$ & $	34.8	$\\
    kitchen-partial	& $	33.8	$ & $	0.0	$ & $	13.1	$ & $	0.0	$ & $	18.9	$ & $	43.9	$\\
    kitchen-mixed	& $	47.5	$ & $	2.5	$ & $	47.2	$ & $	0.0	$ & $	8.1	$ & $	40.8	$\\ \hline

  \end{tabular}
\end{table}

\newpage
\section{Sensitivity Analysis: Max Latent Action}
\label{ab:ma}
\begin{figure}[H]
\centering
\includegraphics[width=\linewidth]{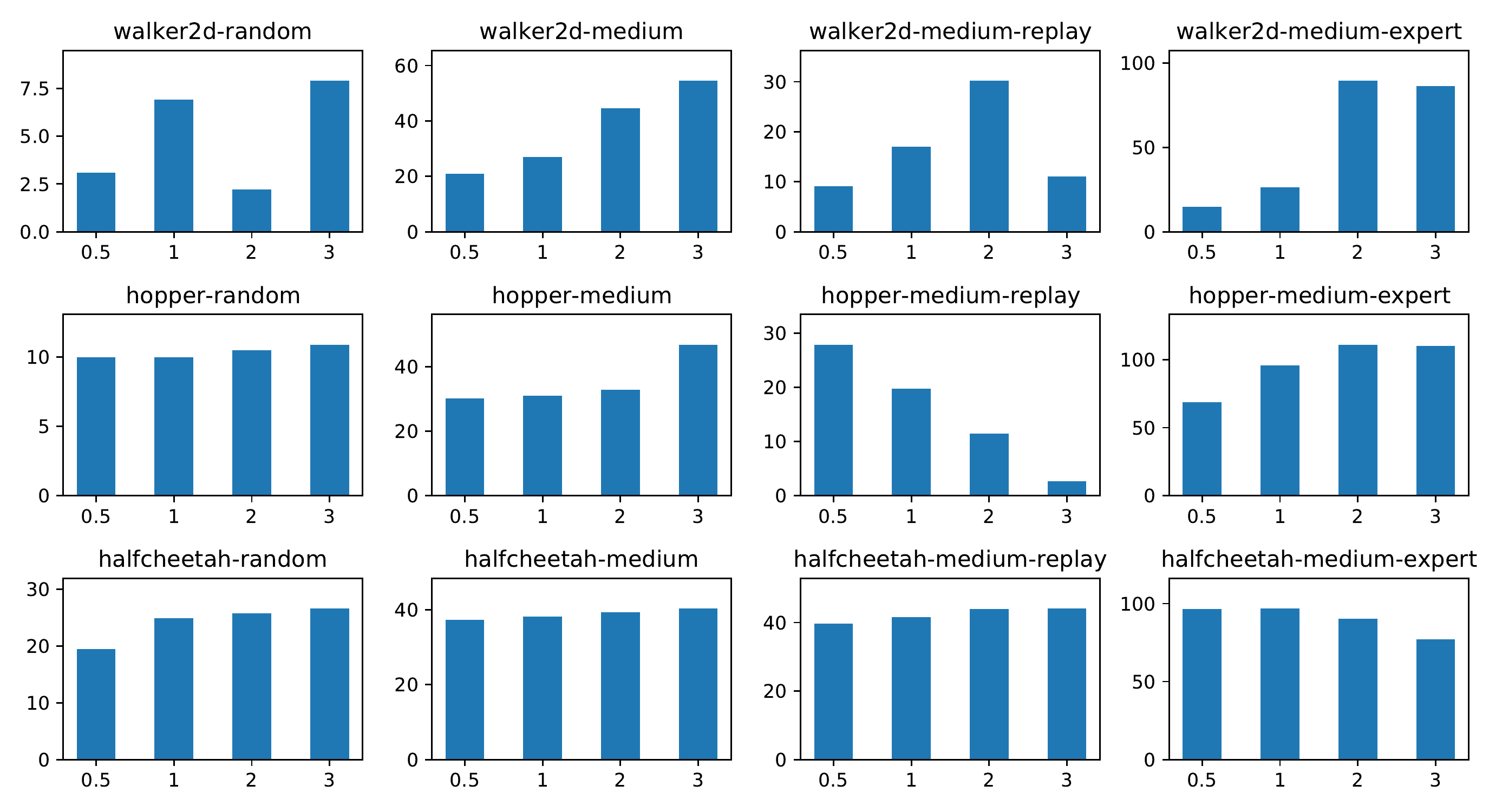}
\caption{Sensitivity analysis on the max latent action for the latent policy: X-axis is the max latent action value. Y-axis is the normalized score.}
\label{fig:ma}
\end{figure}

The max latent action limits the range of output for the latent policy to ensure that the output has a high probability under the latent variable prior of the CVAE. As mentioned in Section \ref{method:latent_policy}, if the output of the latent policy has a high probability under the distribution of the latent variable prior, then the decoded output has a high probability to be within the distribution of the behavior policy. Larger max latent action may result in out-of-distribution actions. On the other hand, smaller max latent action will make the action selection more restrictive. We evaluated the effect of the max latent action from $\{0.5, 1, 2, 3\}$ over the MuJoCo datasets in d4rl as shown in Figure \ref{fig:ma}. In hopper-medium-replay and halfcheetah-medium-expert, $0.5$ works the best. In most cases, $2$ works well. Thus, we use $0.5$ for hopper-medium-replay and halfcheetah-medium-expert and $2$ by default for all the other environments for simplicity. Note that all the experiments for walker2d-random are unstable, thus the comparison across different parameters might not be valuable since we only average across 3 seeds.

\newpage
\section{Ablation Study: Perturbation Layer}
\label{ab:perturbation}

\begin{figure}[H]
\centering
\includegraphics[width=\linewidth]{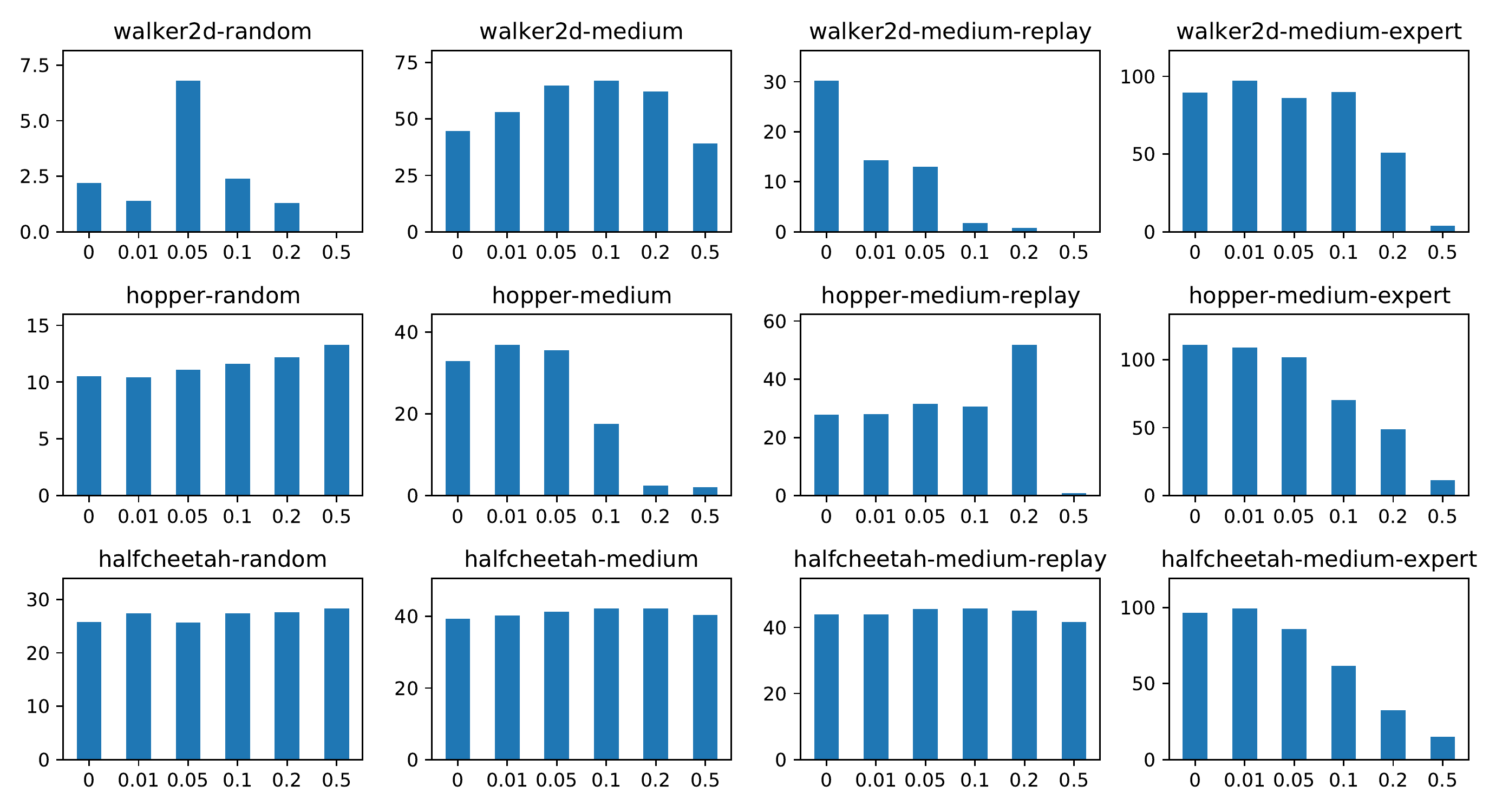}
\caption{Ablation study on the perturbation layer: X-axis is the max perturbation. Y-axis is the normalized score.}
\label{fig:perturbation}
\end{figure}

We provide a full comparison of the perturbation layer on MuJoCo datasets in this section. We summarize the results with max perturbation $\epsilon\in \{0, 0.01, 0.05, 0.1, 0.2, 0.5\}$ in Figure \ref{fig:perturbation}. $\epsilon=0$ is only using the Latent Policy without the perturbation layer. As mentioned above, the walker2d-random experiments are not stable, thus the comparison might not be valuable. In most cases, the addition of the perturbation layer sometimes improves the performance, but not significant. With a large $\epsilon$ higher than a certain value, the performance usually drops. Thus, we make the perturbation layer an optional component in our method.

\newpage
\section{Empirical Analysis on MMD Constraint}
\label{ab:mmd}

To understand the limitation of using sampled MMD constraint to limit out-of-distribution actions, we simulate the MMD loss in different scenarios. In the first experiment, we construct a one-dimensional behavior policy sampling from $N(0,1)$ and an agent policy sampling from $N(0,x)$, where $x$ is a variable. In Figure~\ref{fig:mmd_N} below, we plot the MMD loss for this agent policy with different $x$ as the x-axis with various kernel parameters. Ideally, the loss should be smaller than a threshold for any $x\leq1$ to allow the agent policy to select the best action within the support with a higher probability. However, as shown in Figure~\ref{fig:mmd_N}, this is only roughly satisfied with the Gaussian kernel and large sigma.  Sampled MMD constraint aims to match the entire support of two distributions and could be overly restrictive.

\begin{figure}[H]
\centering
\vspace{-3mm}
\includegraphics[width=0.85\linewidth]{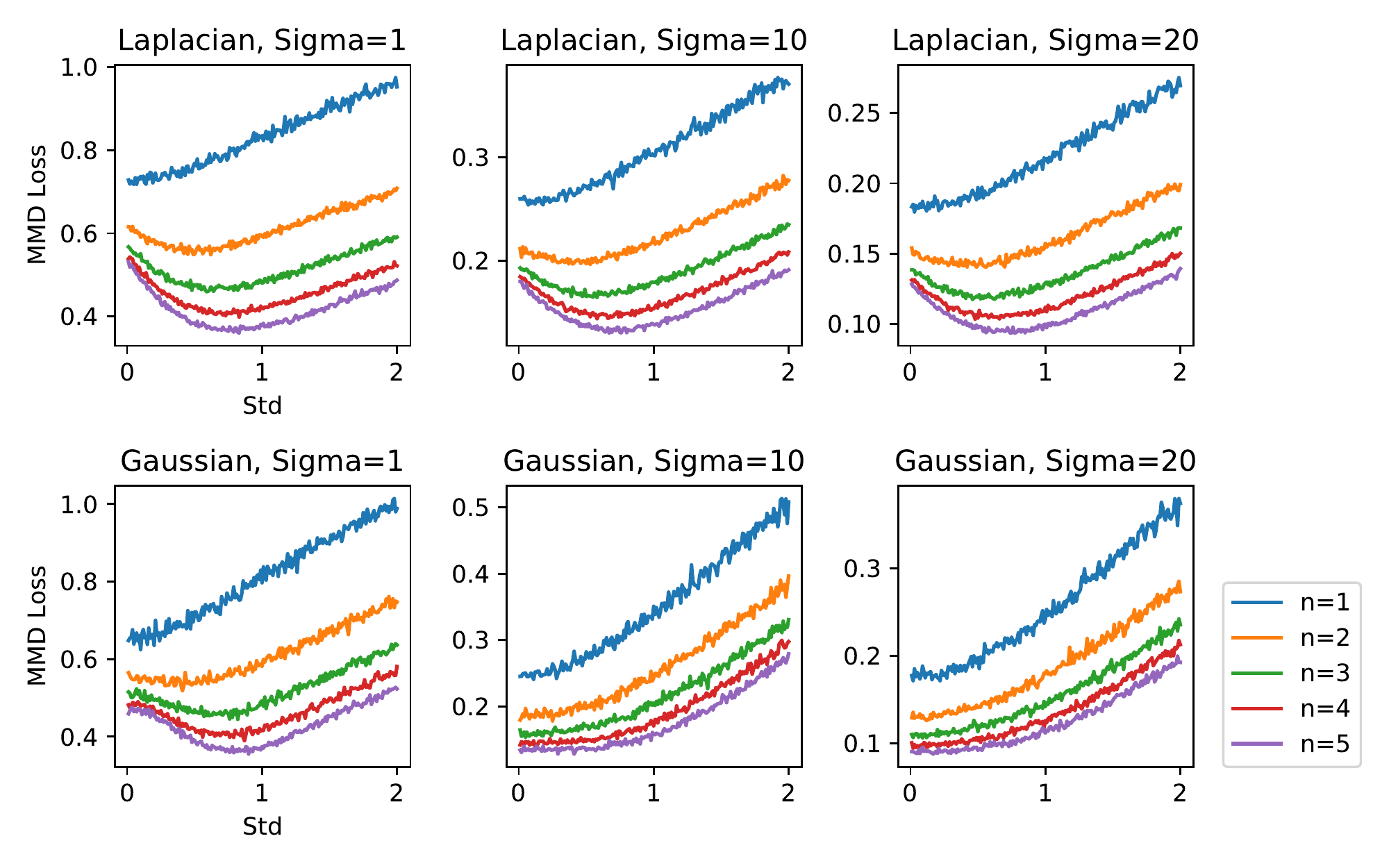}
\vspace{-3mm}
\caption{Simulated MMD loss with N(0,1) as the behavior policy. }
\vspace{-3mm}
\label{fig:mmd_N}
\end{figure}

\begin{figure}[H]
\centering
\vspace{-3mm}
\includegraphics[width=0.85\linewidth]{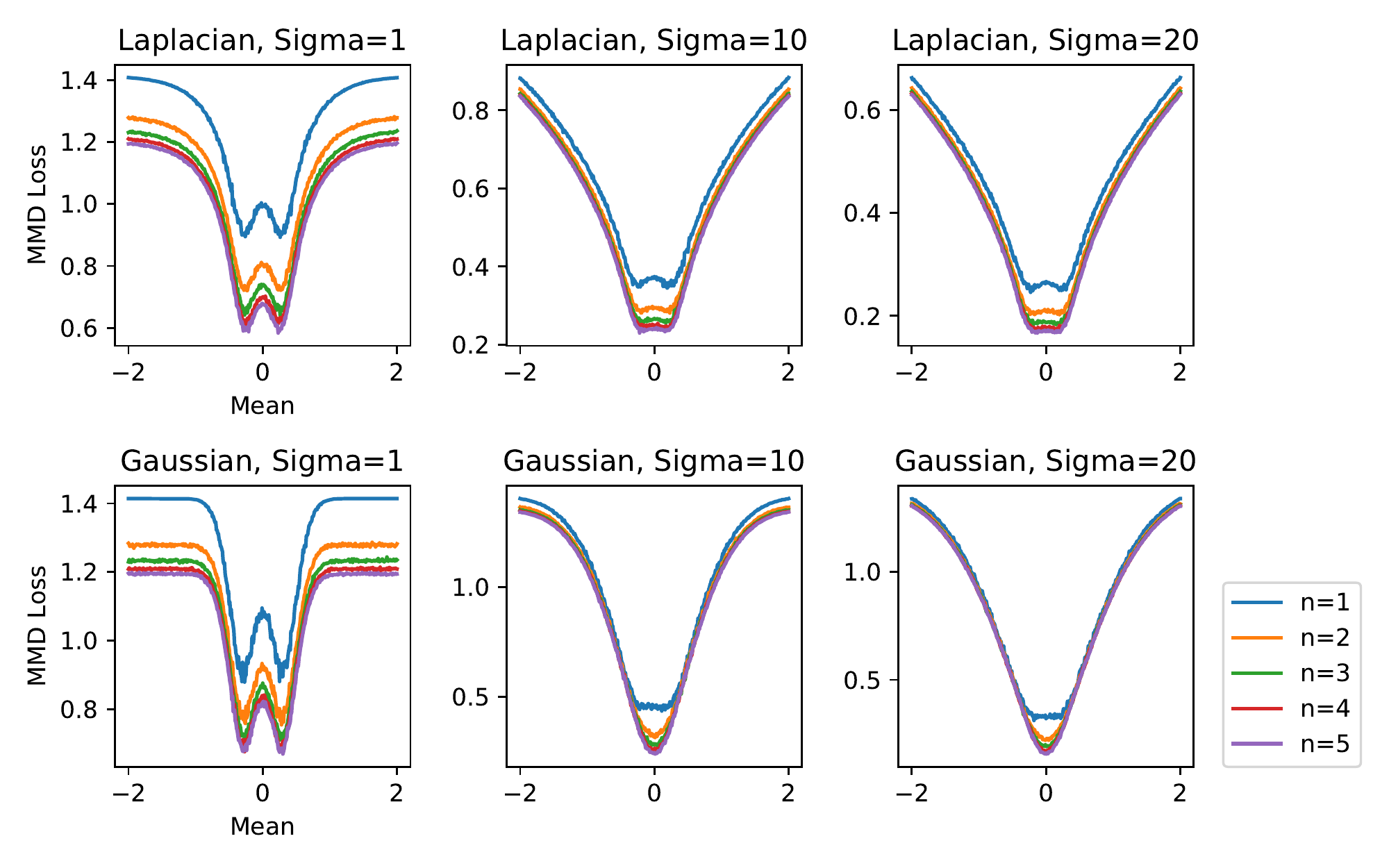}
\vspace{-3mm}
\caption{Simulated MMD loss with N(0,1) as the behavior policy. }
\label{fig:mmd_multimodal}
\end{figure}

In Figure~\ref{fig:mmd_multimodal}, we further demonstrate the limitation of MMD constraint on multimodal distributions. We assume a behavior policy sampling uniformly from $[-2,-1] \bigcup [1,2]$ and an agent policy sampling from $N(x,0.5)$. We vary the mean value $x$ in the x-axis of the figures. In this case, we expect the minimum loss to happen at $x=-1.5$ and $x=1.5$ to prevent out of distribution actions. However, the simulation results show that this is not the case for any of the curves. With large sigma, the minimum MMD loss occurs at $x=0$, which lies in the ``hole'' of the behavior policy distribution.

\newpage
\section{Robot Experiment}
\label{ab:robot}
For the real robot experiments, we use a Sawyer robot equipped with a WSG 32 gripper and a WSG DSA tactile sensor finger. The physical setup involves a cloth with one corner clamped on a fixture. The task is to slide along the cloth as far as possible, ideally until the other corner is reached. 

The movement of the end-effector is constrained to a vertical plane from the fixture. The environment's action space consists of the incremental movement in horizontal (x) and vertical (z) directions for a single time step. The observation space of the environment consists of the tactile sensor readings, end-effector force, and end-effector pose (z position and angle). The observations are thus in the form of a 89-d vector. The action space consists of horizontal and vertical delta position actions. 

The reward and the terminal conditions are designed to encourage large movement in the x-direction without losing the cloth. For each timestep, if the gripper is sliding along the cloth, it receives a reward equal to the horizontal action. This is to encourage faster sliding. However, if the edge is lost from the gripper, the reward will be zero for that timestep and the episode ends. This failure condition is detected based on the gripper width adjustment procedure discussed below. In addition, the maximum episode length is 70 timesteps.

We use a hard-coded procedure to adjust the gripper width. The overall objective is to get clearer tactile readings of the cloth by grasping tightly while allowing the gripper to slide easily without too much friction. The adjustment is based on the coverage and the mean value of the tactile readings as well as the end-effector force readings. When the gripper width is at the minimum value and there is still no tactile reading or force reading, we consider it as a failure and the episode ends.
\end{appendices}

\end{document}